\documentclass{article}

\usepackage{microtype}
\usepackage{graphicx}
\usepackage{subfigure}
\usepackage{booktabs} 
\usepackage{tabularx}
\usepackage{listings}

\usepackage{hyperref}

\usepackage[accepted]{icml2025}

\usepackage{amsmath}
\usepackage{amssymb}
\usepackage{mathtools}
\usepackage{amsthm}

\usepackage[capitalize,noabbrev]{cleveref}

\theoremstyle{plain}

\theoremstyle{definition}

\theoremstyle{remark}

\usepackage[textsize=tiny]{todonotes}
\usepackage{caption}

\icmltitlerunning{Automatic Labelling of PLMs For Generation}

\begin{document}

\twocolumn[
\icmltitle{Automated Neuron Labelling Enables Generative Steering and Interpretability in Protein Language Models}




\begin{icmlauthorlist}
\icmlauthor{Arjun Banerjee}{Berkeley}
\icmlauthor{David Martinez}{Berkeley}
\icmlauthor{Camille Dang}{Berkeley}
\icmlauthor{Ethan Tam}{Berkeley}
\end{icmlauthorlist}

\icmlaffiliation{Berkeley}{Department of Electrical Engineering and Computer Science, University of California, Berkeley, Berkeley, California}

\icmlcorrespondingauthor{Arjun Banerjee}{abaner@berkeley.edu}

\icmlkeywords{Machine Learning, ICML}

\vskip 0.3in
]



\printAffiliationsAndNotice{\icmlEqualContribution} 

\begin{abstract}
Protein language models (PLMs) encode rich biological information, yet their internal neuron representations are poorly understood. We introduce the first automated framework for labeling every neuron in a PLM with biologically grounded natural language descriptions. Unlike prior approaches relying on sparse autoencoders or manual annotation, our method scales to hundreds of thousands of neurons,  raveling individual neurons are selectively sensitive to diverse biochemical and structural properties. We then develop a novel neuron activation-guided steering method to generate proteins with desired traits, enabling  convergence to target biochemical properties like  molecular weight and instability index as well as secondary and tertiary structural motifs—including alpha helices and canonical Zinc Fingers. We finally show that analysis of labeled neurons in different model sizes reveals PLM scaling laws and a structured neuron space distribution.
\end{abstract}

\section{Introduction}
Protein language models (PLMs) have transformed biological sequence modeling, enabling breakthroughs in protein structure prediction, function annotation, and design \cite{rao2021transformer, lin2022evolutionary, Ruffolo_Madani_2024}. Despite the empirical successes of models like ESM-2 \cite{lin2022evolutionary} and ProtTrans \cite{elnaggar2022prottrans}, the internal representations of these models remain opaque, making it difficult to understand how specific features of protein sequences are represented. This lack of interpretability poses barriers to rigorous analysis of model knowledge and hinders the ability for researchers to generate proteins with specific features, which has long been a goal of De Novo Protein design. 

Early studies in PLM interpretability relied largely on heuristic probing, which showed that PLM attention heads encoded structural information about proteins \cite{vig2020bertology, rao2021transformer} and are able to identify functional sites like allosteric residues \cite{kannan2024allosteric, dong2024alloallo}. Going beyond heuristic probing, recent work have leveraged sparse autoencoders (SAEs) to show that  sparse latents in PLMs can capture binding sites, structural motifs, and functional domains that can be used for steering ~\cite{simon2024interplm, adams2025mechanistic, parsan2025matryoshka}. However, SAEs require training a new model on top of the PLM, which can introduce optimization instability, architecture-specific biases, and sensitivity to initialization, which all hinder effectiveness \cite{kantamneni2025sparse, chaudhary2024evaluating, farrell2024applying}.

In parallel, neuron-level labeling has emerged as a promising approach for interpreting model internals, with early methods showing that neurons in vision models capture human-aligned features \cite{bau2017network}. Recent work has extended labelling by using natural language descriptors to GPT-2 \cite{bills2023language} and introducing high-fidelty frameworks that combines exemplar mining, simulator-based scoring, and distillation of high-quality neuron explanations across an entire model \cite{choi2024automatic}. 

\subsection{Contributions}
We apply hidden-unit labeling to PLMs, which enables 2 novel contributions:
\begin{enumerate}
    \item \textbf{Neuron Understanding:} We identify individual neurons that encode broad physicochemical properties—such as charge and hydrophobicity— and distinct structural motifs like zinc fingers or supercoils. We then explore  scaling laws and neuron  representations.
    \item \textbf{Text2Protein Generation:} We leverage LLM-guided neuron queries to generate protein sequences with highly specific features. We show convergence of characteristics (GRAVY, weight, etc.) and both secondary and tertiary structure.
\end{enumerate}

\section{Methods}
At a high level, our problem is to label each neuron in a protein language model with the biophysical feature they are associated with. Then, we seek to use these labels to steer proteins towards features we desire.\footnote{All code is available at: https://github.com/arjun-banerjee/PLMNeuron}

\subsection{Problem Setup}
 More formally, let $\phi: \mathcal{X} \to \mathbb{R}$ denote a scalar-valued neuron feature defined over protein sequences $x \in \mathcal{X}$, extracted from a fixed layer of a pretrained protein language model (PLM). Each sequence $x_i$ is associated with a feature vector $f_i \in \mathcal{F}$ representing structured biological annotations.

Our goal is to produce a natural language description $h \in \mathcal{H}$ that accurately summarizes the conditions under which $\phi(x)$ exhibits strong activation.

Once we obtain a natural language description \( h \in \mathcal{H} \) for each neuron feature \( \phi \), we leverage these interpretable neuron-level labels to guide protein design and analysis. Specifically, given a target biological property \( \tau \in \mathcal{F} \), we identify neurons \( \phi \) whose descriptions \( h \) indicate sensitivity to \( \tau \), and modulate their activations within the PLM to bias outputs toward sequences with enhanced expression of \( \tau \). This enables gradient-free, concept-level interventions on protein sequences, using the neuron basis as a control interface for steering model behavior in alignment with biophysical objectives.

\subsection{Model}

 To achieve this, we create the following pipeline as shown in Figure~\ref{fig:model}:

  \begin{figure}[H]
    \centering
    \includegraphics[width=1\linewidth]{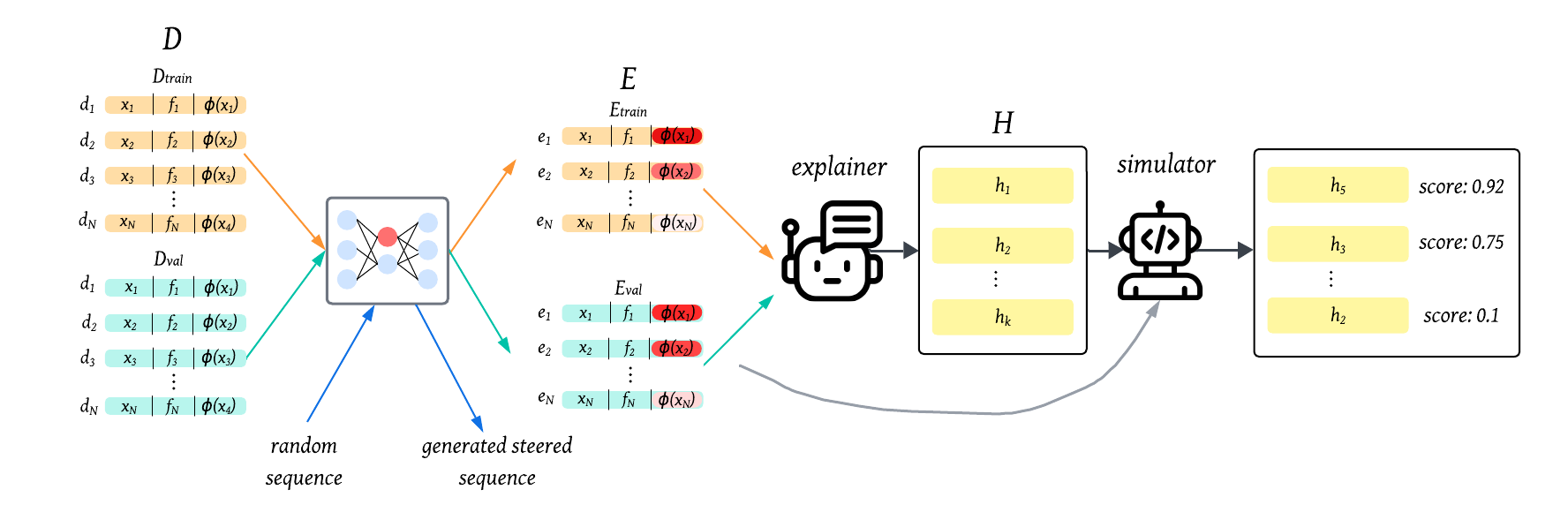}
    \caption{Pipeline for Automatic Neuron Labelling}
    \label{fig:model}
\end{figure}

\begin{enumerate}
    \item \textbf{Dataset Construction:} We create a dataset
    $\mathcal{D} = \{(x_i, f_i, \phi(x_i))\}_{i=1}^n$, $\phi(x_i) \in \mathcal{R}^{+}$ is the normalized activation of neuron $\phi$ on a protein sequence $x_i$.
.
    \item \textbf{Explainer Model:} We create an explainer function $E: \mathcal{D}_{\text{top}} \subset \mathcal{D} \to \mathcal{H}$ which maps a subset of top-activating examples to a hypothesis $h$ in natural language.

    \item \textbf{Simulator Model:} We create a simulator 
    $S: \mathcal{H} \times \mathcal{X} \times \mathcal{F} \to [0, 10]$ which predicts a discretized activation score $\hat{\phi}(x)$ given a description $h$, a sequence $x$, and its features $f$. The goal of this simulator is to find the hypothesis that maximizes the alignment of the hypothesis with the observed activation: $r(h) = \mathrm{Corr} \left( S(h, x_j, f_j), \phi(x_j) \right), \quad \text{for } (x_j, f_j) \in \mathcal{D}_{\text{val}}$, where $\mathrm{Corr}(\cdot, \cdot)$ denotes the Pearson correlation coefficient.

    \item \textbf{Generator:} We aim to generate protein sequences $x \in \mathcal{X}$ that maximize the activation of a subset of neurons $\{z_k\}$ in a fixed layer $\ell$ of a pretrained protein language model $f$. These neurons are selected based on their known association with a desired biological property. Starting from a randomly initialized sequence, we iteratively apply masked inpainting: at each step, we randomly mask a fraction of residues in $x$, run the masked sequence through $f$, and apply an affine intervention $z_k^* = a \cdot z_k(x) + b$ to the selected neuron activations via forward hooks. The modified activations are then propagated through the model to update output logits, and the masked tokens are resampled from the softmax distribution.

\end{enumerate}
\textbf{Dataset Construction}: We initially randomly sample $500,000$ protein sequences under 1024 amino acids from the UniProt Dataset \cite{uniprot2023}, which includes manually labeled qualitative and quantitative descriptors (such as sequence features and functional annotations). We use these descriptors as the features that describe a given protein; we enrich these textual descriptors with additional quantitative data using the \texttt{BioPython} \citep{cock2009biopython} and \texttt{Modlamp} \citep{muller_modlamp_2017} libraries \footnote{Dataset at: https://huggingface.co/datasets/protolyze/plminterp}. A complete list of features and sequence dataset analysis can be found in \autoref{firstappendix}. 

We then place forward hooks at each linear layer of the ESM model and record how much each full protein sequence activates each neuron. We store the sequences that cause the most and least relative activation per neuron. 

\textbf{Explainer Model:} To generate concise natural language explanations for each neuron, we use a prompt-based method that asks a language model to analyze the biological patterns shared by the sequences that most strongly activate that neuron. For each neuron, we collect a list of the $k$ top-activating protein sequences along with their associated quantitative and qualitative features. These are formatted into a structured prompt that instructs the model to holistically generalize the key shared biological traits across the examples. 

We use OpenAI’s GPT-4.1-nano as our explainer model and sample $m$ multiple candidate explanations per neuron using a high-temperature $(T = 0.90$) setting to encourages diverse hypotheses and allow the model to explore multiple plausible generalizations. Each prompt explicitly instructs the model to identify only stable, non-varying features, to avoid redundancy, and to generate a single, concise sentence without introductory phrases. The prompt can be found in \autoref{secondappendix} 

\textbf{Simulator Model:} To evaluate which hypotheses best explain a neuron's behavior, we train a LLM that performs in-context prediction of activation levels. The simulator takes as input a hypothesis, a protein sequence, and its associated biological features, and predicts a discrete activation value from 0 to 10. The hypothesis that best predicts the real activation is considered to be the most accurate. 

We fine-tune Longformer to serve as the simulator due to its ability to handle long context lengths \cite{beltagy2020longformer}. Each input example is formatted as a structured natural language prompt that includes the hypothesis, followed by the amino acid sequence and a dictionary of feature-value pairs. The target is the neuron's normalized activation value, bucketed into one of 11 classes -- which was empirically found to be optimal for analyzing activations.  

We then perform gradient descent on the cross-entropy loss between the model’s predicted class distribution and the ground truth activation bucket. This directly trains the model to use the hypothesis in-context to predict how strongly a neuron will activate on a given input. At evaluation time, we apply the simulator to held-out sequences and compute the Pearson correlation between its predicted scores and the true activations. This correlation score is used to rank hypotheses and select the best explanation for each neuron. The prompts for this can be found in \autoref{secondappendix}. 

\textbf{Generator:} To understand which neurons should be active for a given feature, we query an LLM to assess whether a given label indicates association with the feature. The prompt for this is in \autoref{secondappendix}.

We then follow a similar method to Garcia 2025 \cite{garcia2025interpretingsteeringproteinlanguage}, but we operate on the neuron level and using the ESM tokenizer rather an an encoder/decoder. Beginning with a randomly initialized sequence, we mask a random set of amino acids and forward it through the model to extract hidden representations and identify the activation of the target neuron. An affine transformation ($Ax +B$) is then applied to increase the neuron's activation magnitude, after which the modified hidden state is propagated through the remainder of the model to produce updated output logits. A new set of sampled tokens are sampled from the resulting distribution and decoded via the tokenizer to produce a sequence. This procedure is iterated for a fixed number of steps, and the sequence yielding the highest observed activation is retained. In doing so, we perform a form of activation-guided sequence optimization directly in neuron space.

\section{Generation Results}
We explore our generation mechanism on both characteristic and structural guidance schemes with 2 differently sized models: \texttt{esm2\_t6\_8M} and \texttt{esm2\_t12\_35M}.

\subsection{Characteristic Guidance}
\subsubsection{Single-Characteristic Steering}

To assess characteristic guidance, we activate neurons associated with specific biochemical properties via the previously discussed generation loop. We focus on single-feature steering, in which we select neurons associated with a target descriptor $\tau$ and iteratively steer a randomly initialized sequence to enhance or suppress the corresponding property.

\begin{figure}[H]
    \centering
    \includegraphics[width=1\linewidth]{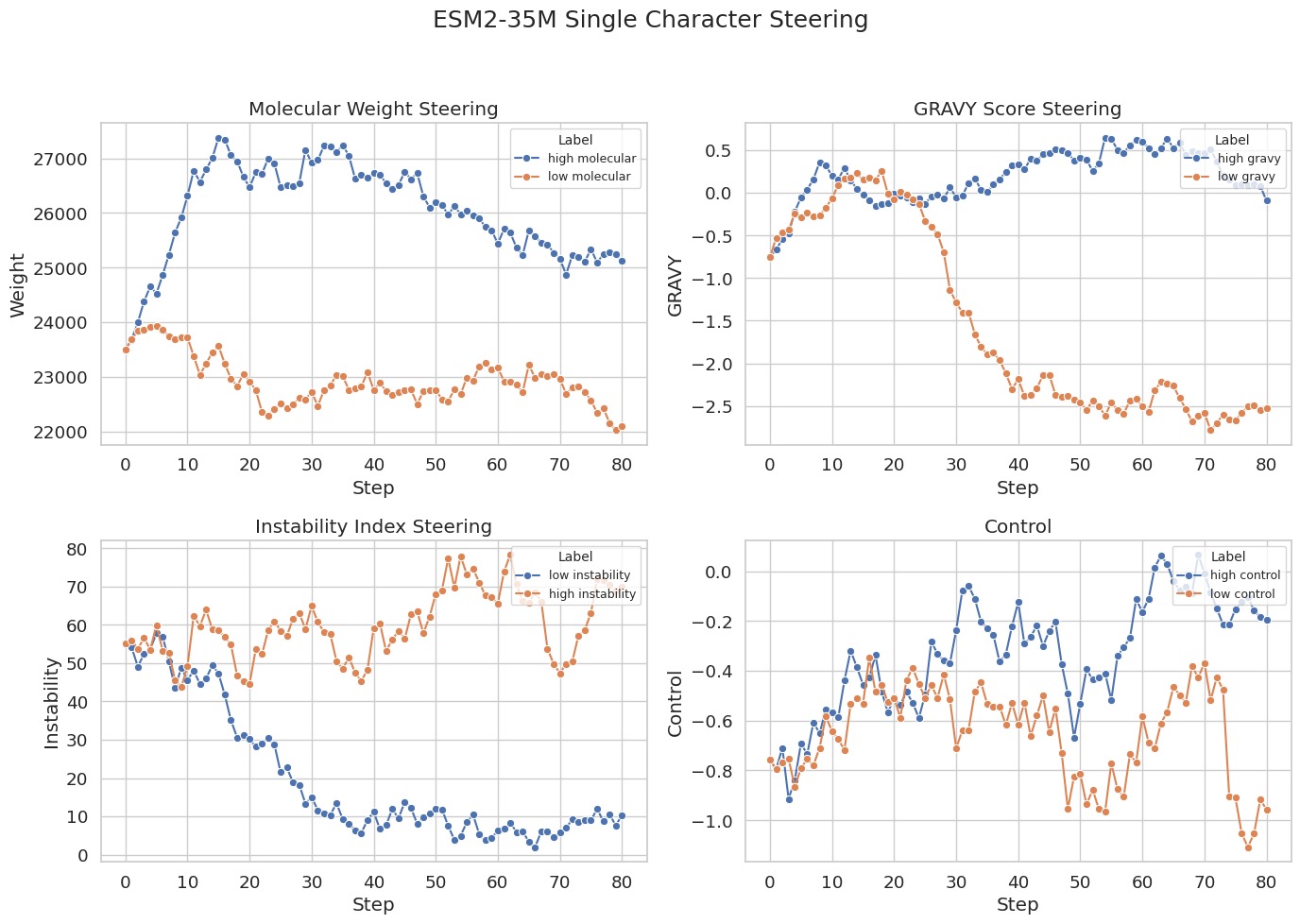}
    \caption{Steering with \texttt{esm2\_t12\_35M}, A = 10 and B = 3 }
    \label{fig:sc3B}
\end{figure}

\begin{figure}[H]
    \centering
    \includegraphics[width=1\linewidth]{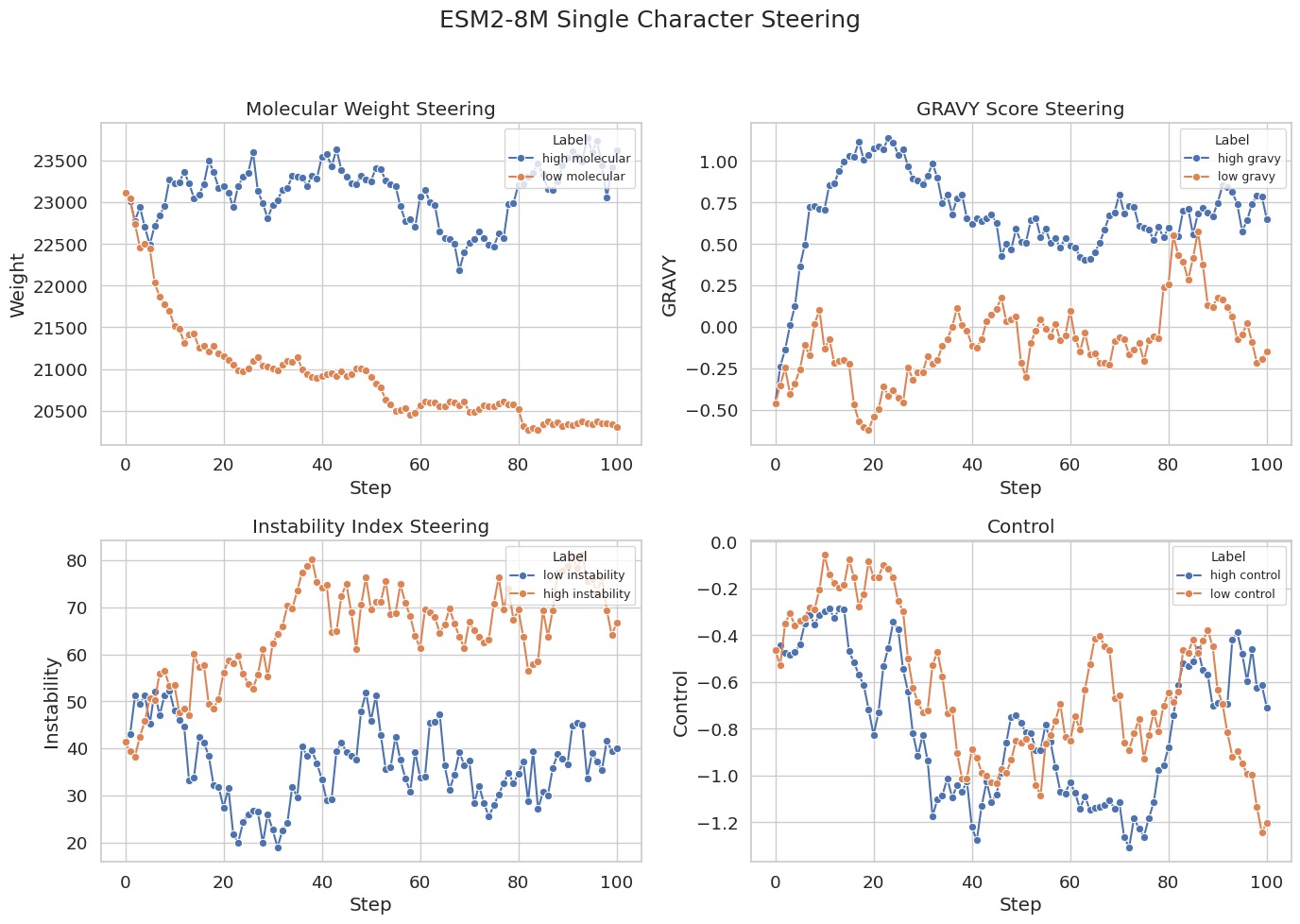}
    \caption{Steering with \texttt{esm2\_t6\_8M}, A = 200 and B = 10}
    \label{fig:sc35M}
\end{figure}
We evaluate the effectiveness of this intervention across three biochemical metrics: molecular weight, GRAVY score, and instability index. For each property, we conduct two experiments---steering toward a ``high'' and ``low'' variant of the feature---using independently annotated neurons identified via semantic search. As seen in Figure~\ref{fig:sc3B} and Figure~\ref{fig:sc35M}, we observe consistent and monotonic divergence between the ``high'' and ``low'' trajectories. For instance, sequences steered toward higher molecular weight exhibit a clear upward trend, while those steered toward lower weight decline generally steadily. Similar patterns emerge for GRAVY score and instability index.

To validate that these changes are due to semantically meaningful interventions, we include a control condition where neurons not associated with any biological property are randomly selected. As expected, the control experiments show no random change in the target metrics in an unsteered manner, confirming that the observed effects are driven by the neuron labels and not by random perturbations.

\subsubsection{Probing Characteristic Representation Space}
\begin{figure}[H]
    \centering
    \includegraphics[width=1\linewidth]{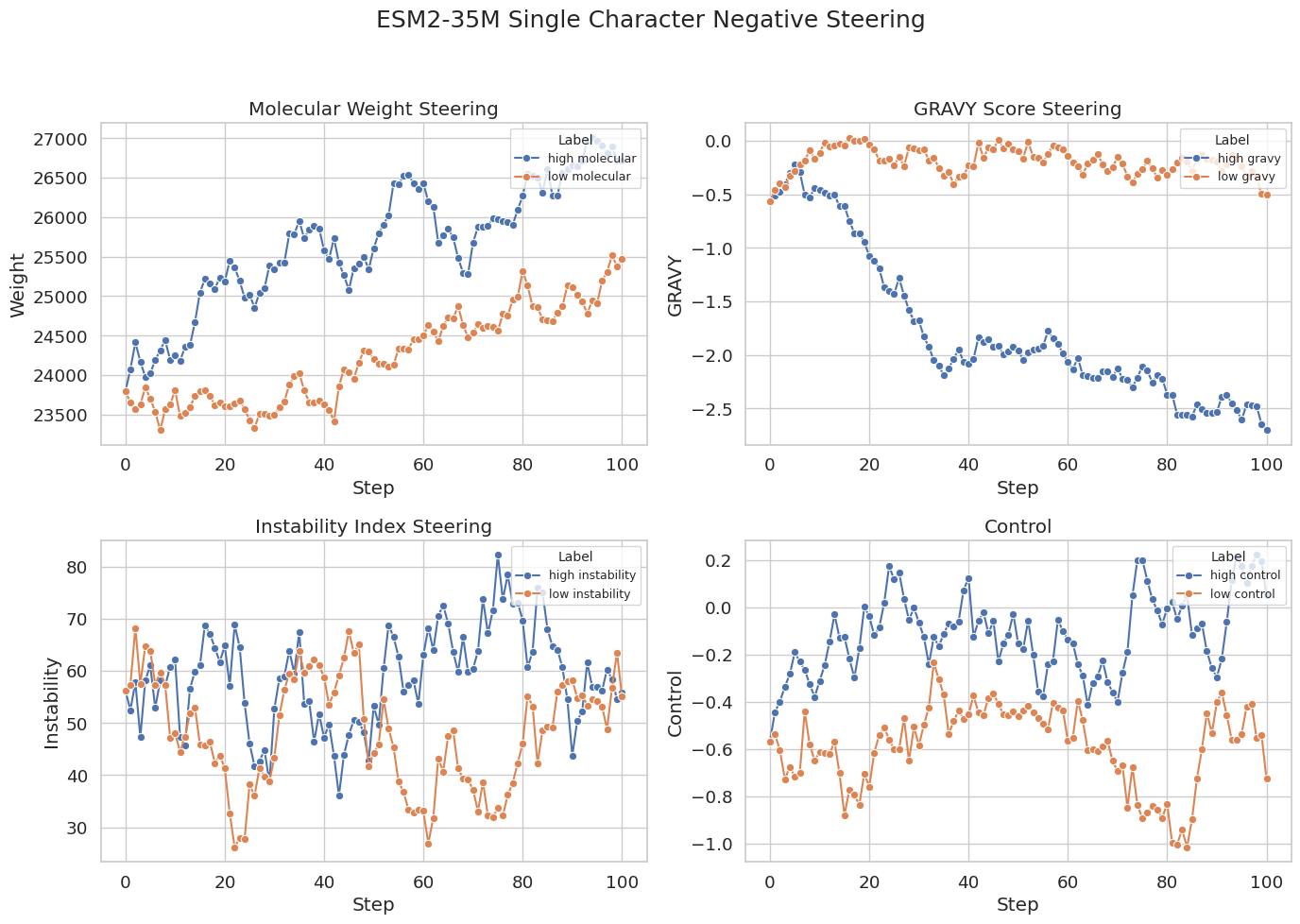}
    \caption{Steering with \texttt{esm2\_t12\_35M}, A = -10 and B = -5}
    \label{fig:negativesteering}
\end{figure}

The existence of neurons labeled with semantically opposing concepts (e.g., 'High $\tau$' vs. 'Low $\tau$') lead to the possible hypothesis that they may encode inverse directions in the model’s representational space. To understand the semantic relationships between such oppositely labeled neurons, we hypothesize that multiplying a 'Low $\tau$' neuron by a negative value should invert its functional effect—producing a result akin to amplifying a 'High $\tau$' neuron. To test this, we introduce \textit{negative steering}, where we multiply the selected neurons by negative $A$ and $B$ values, as shown in Figure~\ref{fig:negativesteering}

We find that the negative steering relationship is highly feature dependent; for instance, in the case of GRAVY score, negative steering led to the complete inverse result of positive steering, while in the case of instability index it led to no conclusive direction changes. Interestingly, in the case of molecular weight, negative steering of low molecular weight and negative steering of high molecular weight \textit{both} increased the weight.

\subsection{Structural Guidance}
\subsubsection{Secondary Structure Guidance}
\begin{figure}[H]
    \centering
    \includegraphics[width=0.75\linewidth]{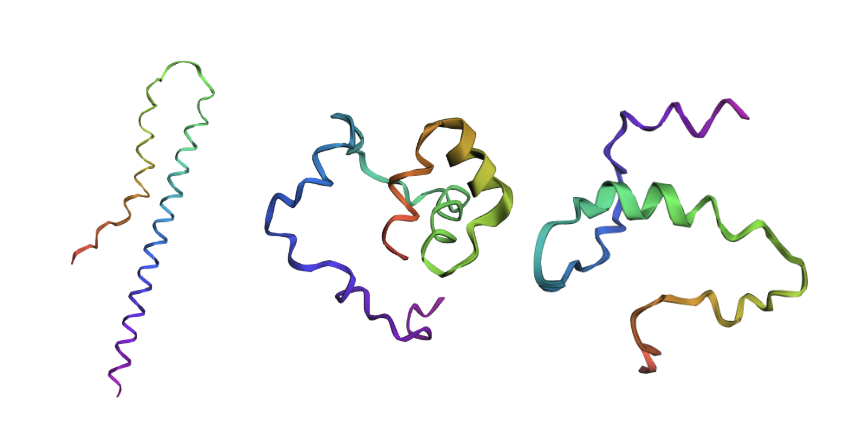}
    \caption{(Left): Visualization of the starting protein; (Middle): Visualization of the alpha helix steered protein; (Right): Visualization of the beta helix steered protein}
    \label{fig:secondary}
\end{figure}

We start by testing the ability of our generator to develop generalized secondary structure by steering towards alpha helices (semantically searching "alpha") and beta strands (semantically searching "beta") starting from a neutral sequence of 75 D-configuration amino acids, respectively. The results of these generations can be seen in Figure~\ref{fig:secondary}

\begin{figure}[H]
    \centering
    \includegraphics[width=1\linewidth]{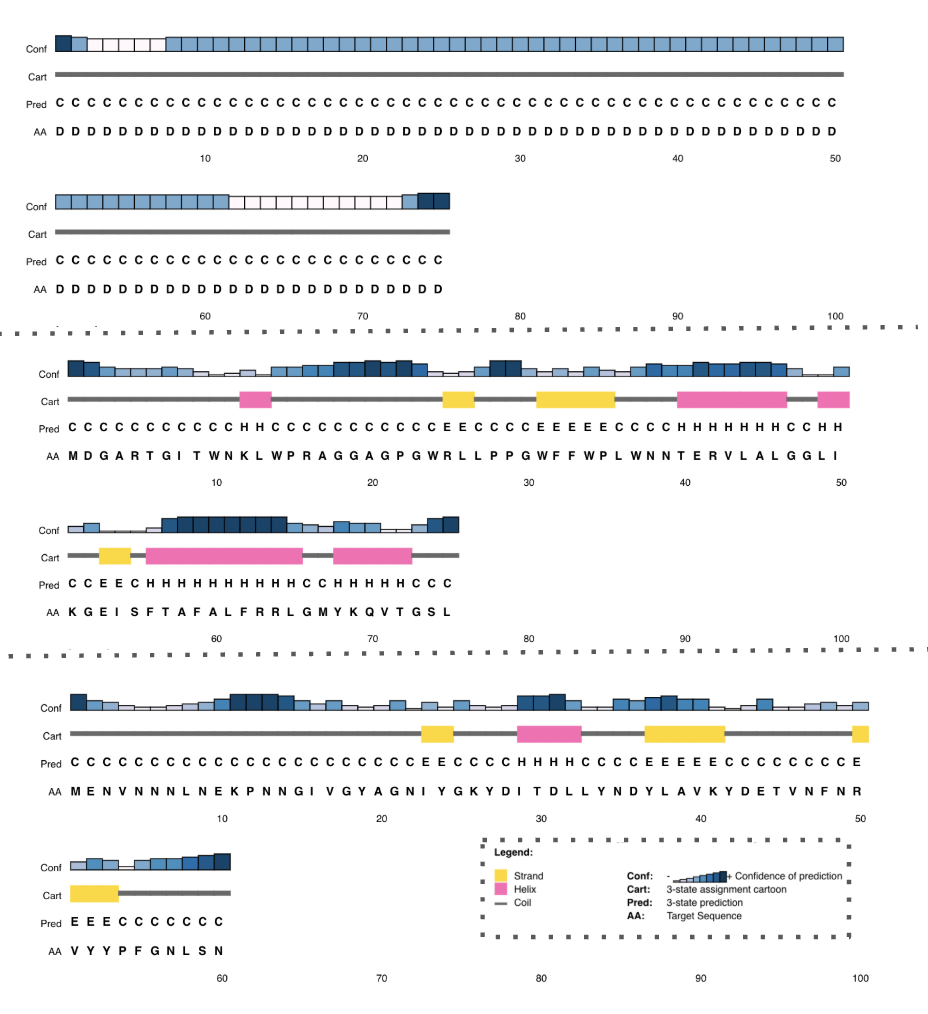}
    \caption{PsiPred outputs of the neutral sequence (top), the alpha steered sequence (middle), and the beta steered seqeunce (bottom)}
    \label{fig:psipred}
\end{figure}

To label the regions of the generated steers, the sequences were fed into PsiPred \cite{mcguffin2000psipred} and their resultant domains were annotated with C (to denote coils), S (to denote beta strands), or H (to denote alpha helices). As depicted in Figure~\ref{fig:psipred}, the alpha steered model contains 5 distinct alpha helix domains spanning 34\% of the protein, while the beta steered model contains 4 distinct beta strand domains. Both proteins also contain the opposing secondary structure motifs (there are 3 beta strand domains in the alpha protein and 1 alpha helix in the beta domain); this result is not surprising, as there are many neurons that are labelled as controlling both alpha and beta secondary structures. 

While these results demonstrate the generators ability to converge on the desired secondary structures, it is important to note that these structures are likely not actualize in the laboratory. The alpha and beta steered proteins have a $\Delta G$ value of $99.64$ and $113.05$ kCal/mol, respectively, which is far above the typical 5-15 kCal/mol range \cite{Ahmad2022Protein}. A more complete analysis of the generated proteins can be found in \autoref{thirdappendix}, and a discussion of how to improve viability can be found in the Future Works section. 

\subsubsection{Tertiary Structure Guidance}
\begin{figure}[H]
    \centering
    \includegraphics[width=0.5\linewidth]{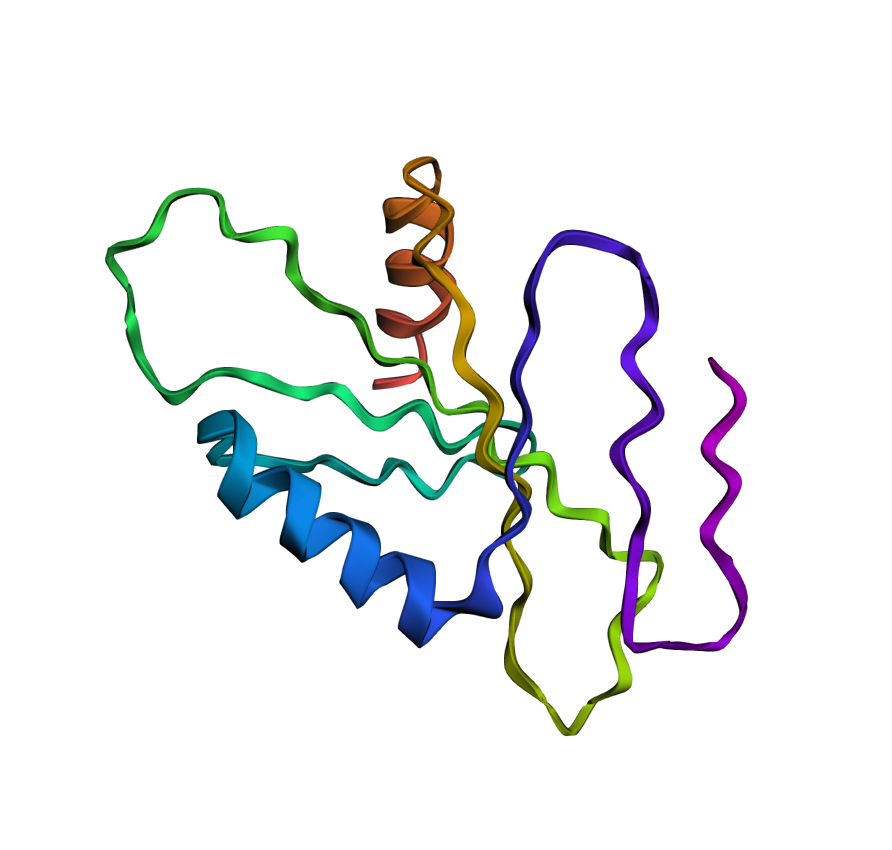}
    \caption{Visualization of a Generated Zinc Finger Motif}
    \label{fig:Zinc}
\end{figure}

To test the ability of our generator to develop tertiary structure, we generate proteins steered towards neurons containing "zinc finger." We were able to observe generations with this characteristic: in Figure~\ref{fig:Zinc}, we display a generated protein with a canonical C2H2 zinc finger motif, characterized by the pattern Cys-X(2–4)-Cys-X(12)-His-X(3–5)-His.

Of the 100 sequences we generated, this was the only fully correct sequences we observed; however, many other sequences contained near hits to Zinc Finger characteristics. One protein generated a "HCCCCACACF" subsequence, which is very similar to the RING finger domain. Another sequence contained a C-X3-C-X20-C-X4-C domain, which resembles a C4 motif. 

We predict that using this approach on a model with more hidden units, like the 15 billion variant, would yield more rich structural information which would enable more accurate generations. However, it is important to note that this protein is likely not able to viably fold given it's $\Delta G$ value of $172.49$; more about this is discussed in \autoref{thirdappendix}.

\section{Interpretability Results}
We label all the neurons of 3 differently sized models: \texttt{esm2\_t36\_8B\_UR50D},  \texttt{esm2\_t12\_8B\_UR50D}, and \texttt{esm2\_t36\_8B\_UR50D}. A random selection of neuron labels across the 12 layers of \texttt{esm2\_t12\_8B\_UR50D} can be seen in Table~\ref{tab:neuron-descriptions}. \footnote{All labels available at https://huggingface.co/protolyze/datasets}
\begin{table}[h]
\centering
\begin{tabularx}{\columnwidth}{|l|X|}
\hline
\textbf{Neuron} & \textbf{Description} \\
\hline
(0, 160) & Strongly activates for secreted proteins with low to negative GRAVY scores. \\
(1, 323) & Strongly activates for flagellin proteins involved in bacterial flagellum structure. \\
(4, 204) & Strongly activates for proteins with high charge at pH 7 and a significant fraction of beta-sheet structure. \\
(7, 467) & Strongly activates for proteins with tryptophan synthase activity and negative gravy scores. \\
(9, 437) & Strongly activates for proteins with a specific role in DNA replication initiation and regulation. \\
(11, 473) & Strongly activates for chloroplastic proteins involved in RNA binding and processing. \\
\hline
\end{tabularx}
\caption{Neuron-level activation descriptions. Each neuron is identified by an ordered pair \((\ell, n)\), where \(\ell\) is the layer index and \(n\) is the neuron index within that layer.}
\label{tab:neuron-descriptions}
\end{table}

\subsubsection{Toward a PLM Scaling Law}
\begin{figure}[H]
    \centering
    \includegraphics[width=1\linewidth]{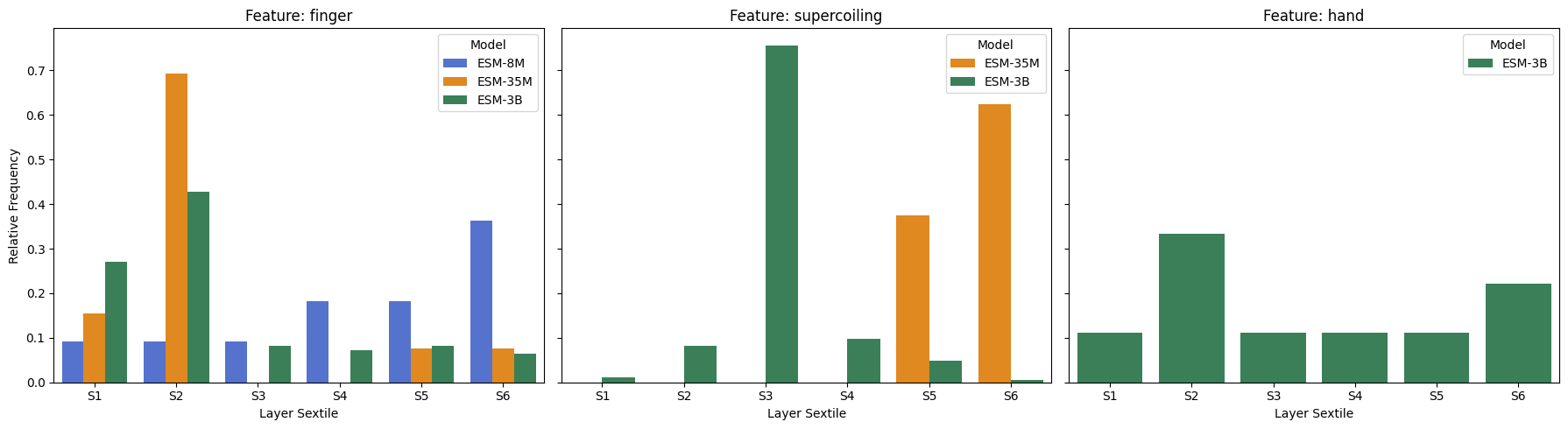}
    \caption{Relative occurance of structural features for each ESM variant labeled by layer sextile. Certain structural features fail to be captured by smaller models. Sextiles were used to ensure accurate relative comparisons between models, as the $8$M variant is 6 layers and all other model's layers are divisible by 6.}
    \label{fig:fine-grain}
\end{figure}

Preliminary analysis of the labels suggested that bigger models are able to capture more fine-grained protein features; for instance, Figure~\ref{fig:fine-grain} demonstrates that the 3B model is able to capture structural details \textit{hands, supercoiling,} and \textit{fingers}, whilst the 8M model is only able to capture \textit{fingers}. As such, we hypothesize that bigger PLMs are able to capture more niche elements of a proteins composition rather than just adding more parameters that represent generic features. 

Another set of phenomena we observed is the late-specialization of functional features in smaller models, possibly attributed to information bottlenecks. In other words, early feed-forward layers in smaller models function as universal encoders, while the last layers are largely discriminative on the set of learned embeddings from earlier layers. In layers S1–S5, ESM-8M exhibits only a trickle of function-related signal; instead, these layers are likely devoted to extracting generalizable sequence patterns—basic amino-acid correlations, local motifs, and low-level statistical features—that will be broadly useful downstream. We observe the functionality shift from encoding to discriminating features in an incremental manner by the roughly linear increase in neuron-specific attribution. In addition, observe  how increasing the parameter size  leads to greater frequency of feature-attributed neurons, suggesting that the model shifts from an encoder-discriminator model to a hierarchal model which expends more layers to learn richer and finer-grain features, which is supported by the emergence of hand features in the ESM-3B model in Figure~\ref{fig:fine-grain}. In smaller models, this is not feasible because the tight information bottleneck in early layers collapsed diverse feature information into a low‐dimensional code, leaving insufficient representational capacity to carve out the fine‐grained detectors necessary for richer features like hands.

We speculate that this behavior stems from the limited parameter budget, which forces the model to spread feature attribution broadly across neurons and inhibits clear specialization. Consequently, early layers operate as a general‐purpose encoder: their activations simultaneously support the original PLM training objective of masked‐amino‐acid prediction while enabling implicit downstream functional protein characteristics to form. Unraveling how information becomes entangled across neuron populations—and how this entanglement influences specialization—remains an important avenue for future investigation.

\subsubsection{Feature Locations}
Table~\ref{tab:neuron-descriptions} suggests that lower-layer neurons capture local biochemical features like charge and GRAVY score (e.g., neuron (0, 160)). Middle layers begin to detect more complex patterns such as secondary structure and domain-like segments (e.g., neurons (4, 204) and (7, 467)), while higher layers abstract to functional roles like DNA replication or RNA processing (e.g., neurons (9, 437) and (11, 473)). This progression seems to reflect the typical hierarchical abstraction in protein language models. 

\begin{figure}[H]
    \centering
    \includegraphics[width=1\linewidth]{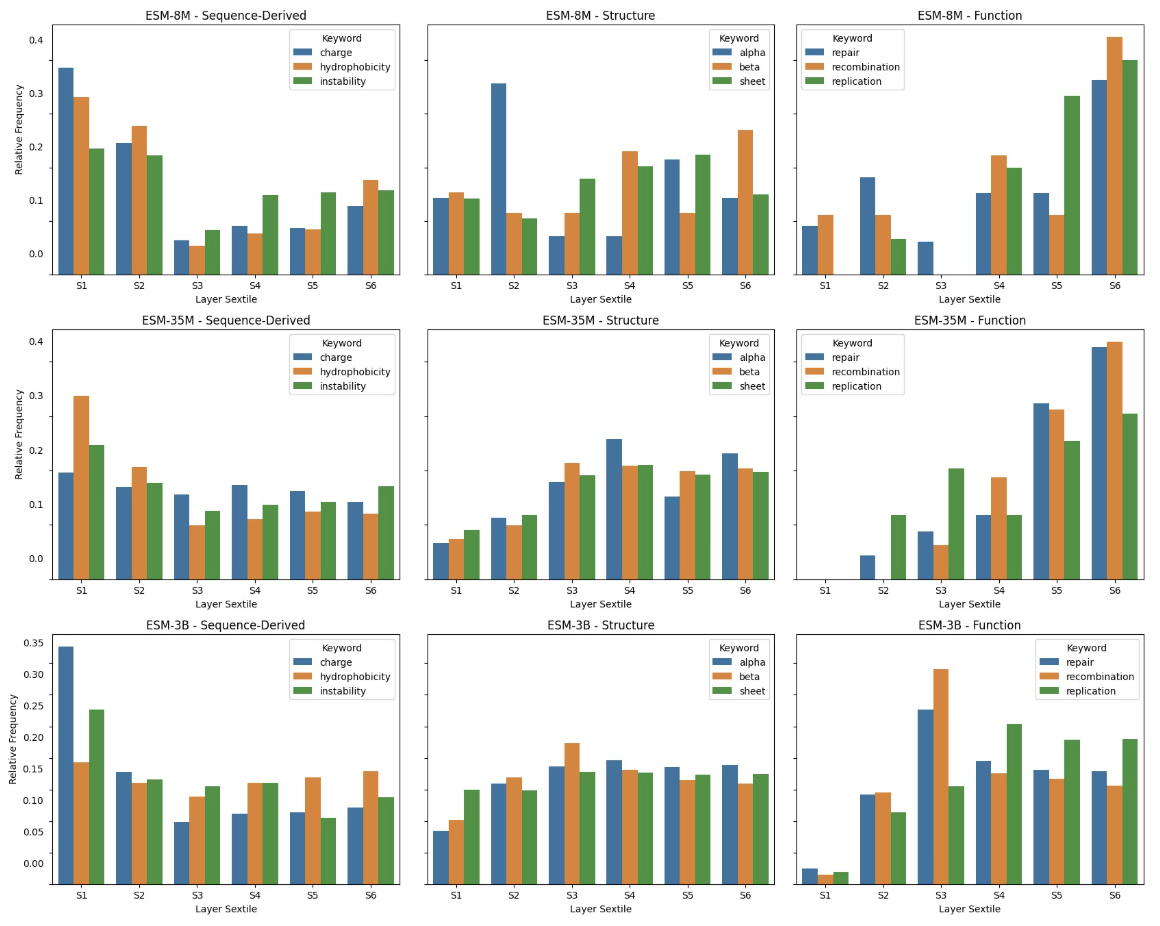}
    
    \caption{Relative occurance of each property (functional, structural, sequence derived) for each ESM variant labeled by layer sextile. Sextiles were used to ensure accurate relative comparisons between models, as the $8$M variant is 6 layers and all other model's layers are divisible by 6.}
    \label{fig:Locations}
\end{figure}

To confirm this hypothesis, we sample 3 characteristics associated with functional roles (\textit{repair}, \textit{recombination}, and \textit{replication}), 3 characteristics associated with structural roles (\textit{sheet}, \textit{alpha}, and \textit{beta}) and 3 characteristics associated with sequence-derived properties (\textit{charge}, \textit{hydrophobicity}, \textit{instability}). We semantically search all the neurons associated with this keyword, and then plot their relative distributions in Figure~\ref{fig:Locations} for each of the model variants. 

The figure above demonstrates how different categories of protein features, from sequence-derived to structural to functional features, are encoded in layers of the PLM.
All model sizes demonstrate some degree of hierarchical encoding. Starting with local biochemical properties / sequence-derived features (charge, hydrophobicity, instability), we can see that its representation peaks in lower layers of the models, and drops in higher sextiles, which is consistent with low-level local feature encoding in earlier layers. This aligns with the intuition that local-level features such as biochemical properties can often be inferred with just 

Structural properties, as shown in the middle column, peak slightly in the middle layers, but persist through the later layers, which reflect an increasing context window that is required to infer secondary and tertiary structural patterns, which often span multiple residues but not the full protein. 
And on the rightmost column, we can see that functional features have the strongest concentration in later sextiles, particularly in the ESM-8M and ESM-35M models, showing that functional properties require global context and tend to emerge in deeper layers of the network.
This progression reflects the pattern typical of language models where shallow layers tend to encode local features, and as layer depth increases, 

In ESM-8M, we can see that the model's layers appear to have a more distinct specialization, such as the sequence-derived features sharply peaking at the first sextile and then quickly falling off after, or functional-related neurons concentrated in the final sextile, which shows that the model delays global / semantic processing until the end.

Meanwhile, as the model size increases, the distributions become more distributed, 
For example, functional features show up in earlier sextiles in ESM-3B as compared to the other two smaller models, indicating greater distribution across the model.

\section{Looking Forwards}

\subsection{Limitations and Future Work}\label{fw}

\textbf{LLM Oversimplification:} Despite providing comprehensive prompts and rich feature sets, the explainer model (GPT-4.1-nano) may still generate oversimplified or overly generic explanations, reflecting limitations in the model’s ability to synthesize complex biological patterns. Additionally, the model might also hallucinate features due to high temperatures, despite the attempts of the simulator to mitigate this. To assess neuron description quality  in the future, it would be useful to generate an evaluation rubric and benchmark for descriptions and then have humans assess the quality of said labels. 

\textbf{Structural Viability:} While our generator can produce proteins exhibiting desired characteristics, it does not guarantee that the resulting sequences will fold into viable, stable structures. To address this, we aim to incorporate a reinforcement learning module into our loop where a structure prediction model (e.g., AlphaFold2 or ESMFold) acts as an oracle to evaluate the foldability of generated sequences. The feedback from this model serves as a reward signal, encouraging the generator to produce sequences that are not only functionally relevant but also structurally plausible. Future work could compare generated sequences to their nearest real neighbors and incorporate $\Delta$ G scores for a more robust assessment.

\textbf{Expand Labels:} Currently, we only label 3 different models and use one LLM due to compute limitations. We plan to extend neuron labeling beyond the three models used in this work to the full suite of ESM models, covering a broader range of architectures and sizes. Additionally, we aim to use multiple large language models and diverse prompting strategies to generate more robust and varied natural language labels. This will help reduce bias from any single LLM and better capture the functional diversity of neurons across models.

\textbf{Model Pruning:} We seek to explore pruning neurons based on their assigned labels, focusing on those with low specificity, high redundancy, or minimal activation. By removing such neurons, we aim to reduce model size and inference cost while retaining the core functional capacity of the model. This approach can support interpretable model compression and shed light on which neurons are essential for biological representation. We hope this will generate a smaller condense variant of ESM with the same performance. 

\subsection{Conclusion}

We introduce a comprehensive framework for interpreting individual neurons in protein language models through natural language explanations, simulation-based validation, and generative steering. By leveraging both qualitative annotations from UniProtKB and computed biochemical features, we provide a scalable and biologically grounded method for neuron labeling in ESM2. Our explainer model generates concise hypotheses that capture the biological patterns associated with each neuron, while our simulator quantitatively evaluates the quality of these hypotheses through in-context prediction. We demonstrate that these labeled neurons can be used to steer sequence generation toward targeted biophysical properties and structural motifs, enabling new modes of interaction with pretrained protein models. We lastly analyze the labels, suggesting elementary scaling laws and analyze feature locations.

Together, these contributions establish a foundation for mechanistically understanding and manipulating PLMs at the neuron level. Future work can further integrate structural validation, extend to other protein models such as AlphaFold and RoseTTAFold, and explore higher-order compositional representations.

Additionally, our generative steering framework lays the groundwork for simultaneously modulating multiple biological properties within a single protein sequence. This multi-objective control could be especially valuable in therapeutic protein design, where optimizing for multiple criteria such as stability and efficacy is essential. By enabling fine-grained, neuron-level manipulation of pre-trained models, we move closer to controllable and interpretable protein generation pipelines that align with practical needs in drug discovery and synthetic biology.

\section*{Impact Statement}
This paper presents work whose goal is to advance the field of 
Machine Learning. There are many potential societal consequences 
of our work, none which we feel must be specifically highlighted here.

\section*{Acknowledgements}
The authors thank Anant Sahai, Naman Jain, and Qiyang Li for feedback in the early stages of the project, and for insightfully teaching the class that made this possible. We are also grateful to Kiran Suresh, Amos You, and Eshaan Moorjani for introducing us to neuron labeling work and helpful discussions regarding scaling.

\section*{Code and Data Availability Statement}
All code can be found publicly on Github at https://github.com/arjun-banerjee/PLMNeuron. All datasets used and labels generated can be found at https://huggingface.co/protolyze/datasets. 

\bibliography{example_paper}
\bibliographystyle{icml2025}

\newpage
\appendix
\onecolumn
\section{Dataset}\label{firstappendix}

\begin{figure}[H]
    \centering
    \includegraphics[width=1\linewidth]{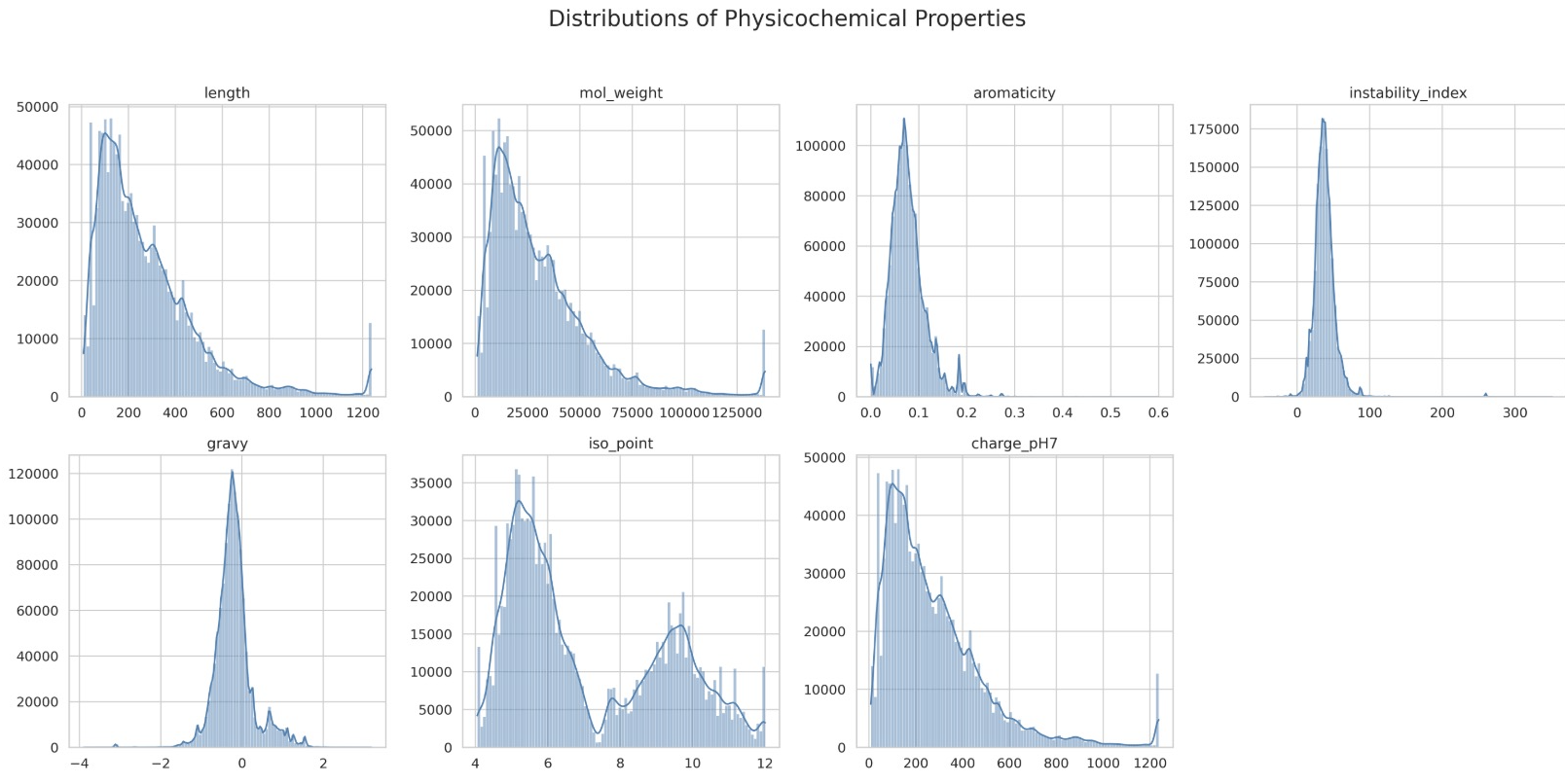}
    \caption{Distribution of features on label dataset}
    \label{fig:datasetds}
\end{figure}

We used 500{,}000 sequences along with qualitative annotations from the UniProtKB dataset~\cite{uniprot2023}. We used the following features: subcellular location, Gene Ontology (biological process and molecular function), catalytic activity, pathway, enzyme commission (EC) number, disruption phenotype, induction, domain, and functional descriptions. In addition, we computed quantitative biochemical features using the \texttt{BioPython} package, including sequence length, molecular weight, isoelectric point, aromaticity, instability index, GRAVY score, and predicted secondary structure fractions (helix, turn, sheet). We also used the \texttt{modlAMP} package to calculate charge at pH 7, Boman index, aliphatic index, and hydrophobic moment. To understand the distribution of the dataset we sampled from, refer to Figure~\ref{fig:datasetds}.

\section{Prompts}\label{secondappendix}
We experimented with a variety of explainer system prompts with varying levels of directions and examples. The system prompt that generated the best results was:
\lstset{
  breaklines=true,
  breakatwhitespace=false,
  basicstyle=\ttfamily\small,
  frame=single
}
\begin{lstlisting}
You are an AI researcher investigating a specific neuron inside a protein language model. Your task is to describe the biological features of protein sequences that cause the neuron to strongly activate. Your goal is to generalize the neuron label by finding patterns in the features across all examples provided.

You will be given the following information:

<protein_sequences>
{{PROTEIN_SEQUENCES}}
</protein_sequences>

<biological_features>
{{BIOLOGICAL_FEATURES}}
</biological_features>

<activation_values>
{{ACTIVATION_VALUES}}
</activation_values>

Analyze the data provided:
1. Examine the protein sequences and their corresponding biological features.
2. Pay attention to the activation values for each sequence.
3. Look for common patterns or characteristics among sequences with high activation values.
4. Consider your knowledge of biology to interpret the significance of these patterns.

Formulate your description:
1. Identify the most important 1-2 features that consistently appear in sequences with high activation.
2. Focus on features that are common across most or all high-activation sequences.
3. Disregard features that vary significantly among the examples.
4. Create a concise, one-sentence description that captures the essence of what causes the neuron to strongly activate.

Output your final description inside <neuron_description> tags. Ensure your description:
- Is limited to one sentence
- Uses as few words as possible
- Directly states the relevant features without introductory phrases
- Describes only consistent patterns across the provided examples

Example high-quality responses:
"Strongly activates for sequences of membrane proteins involved in transmembrane transport processes."
"Strongly activates for proteins with negative gravy scores"
"Strongly activates for glycoproteins involved in cellular structural functions"

\end{lstlisting}

Then, the input prompt was:

\lstset{
  breaklines=true,
  breakatwhitespace=false,
  basicstyle=\ttfamily\small,
  frame=single
}
\begin{lstlisting}
You will be given a list of DNA or protein sequences and their associated biological features where a neuron strongly activates. Your task is to summarize the shared biological features among these sequences in one concise sentence.

Here is the list of sequences and their associated features:

<sequences_and_features>
{{SEQUENCES_AND_FEATURES}}
</sequences_and_features>

To complete this task, follow these steps:

1. Carefully read through all the sequences and their associated biological features.
2. Identify common themes or patterns in the biological features across the sequences.
3. Focus on the most prominent and frequently occurring features.
4. Synthesize these common features into a single, concise statement.

Your summary should capture the essence of the shared biological features using the fewest words possible while still conveying the key information.

Provide your summary in the following format:
<summary>
[Your one-sentence summary of shared biological features]
</summary>

Remember, brevity is crucial. Aim to use no more words than absolutely necessary to accurately convey the shared biological features.\end{lstlisting}

The simulator model was trained using the following prompt:

\begin{lstlisting}
    Task: Predict activation 0 - 10. ONLY ANSWER WITH A NUMBER
    Neuron: {row["neuron_id"]}
    Description: {hypo}
    Sequence: {seq}
    Features: {comp}
    ONLY ANSWER WITH A NUMBER BETWEEN 0 AND 10.         
\end{lstlisting}

Where the appropriate values replace the elements in the brackets. Inference was then done with this prompt as well. 

Then, the prompt for neuron selection was simply:
\begin{lstlisting}
    Answer only with a True or False. A neuron described as {neuron} be useful in trying to generate a protein with the following characteristic: {characteristic}?

    For example: 
    Prompt: "Answer only with a True or False. A neuron described as "Encodes information about Zinc Fingers" be useful in trying to generate a protein with the following characteristic: "Alpha-Sheet"?", Answer: False
    Prompt: "Answer only with a True or False. A neuron described as "Associated with high hydrophobicity" be useful in trying to generate a protein with the following characteristic: "Increasing hydrophobicity"?", Answer: True
\end{lstlisting}
\section{Analysis of Generated Proteins}\label{thirdappendix}

To understand how likely our steering loop is to generate proteins not in the distribution of proteins the neurons were labeled with, we plot the characteristics of the generated proteins against the distribution of the dataset we used to generate labels.

  \begin{figure}[H]
    \centering
    \includegraphics[width=1\linewidth]{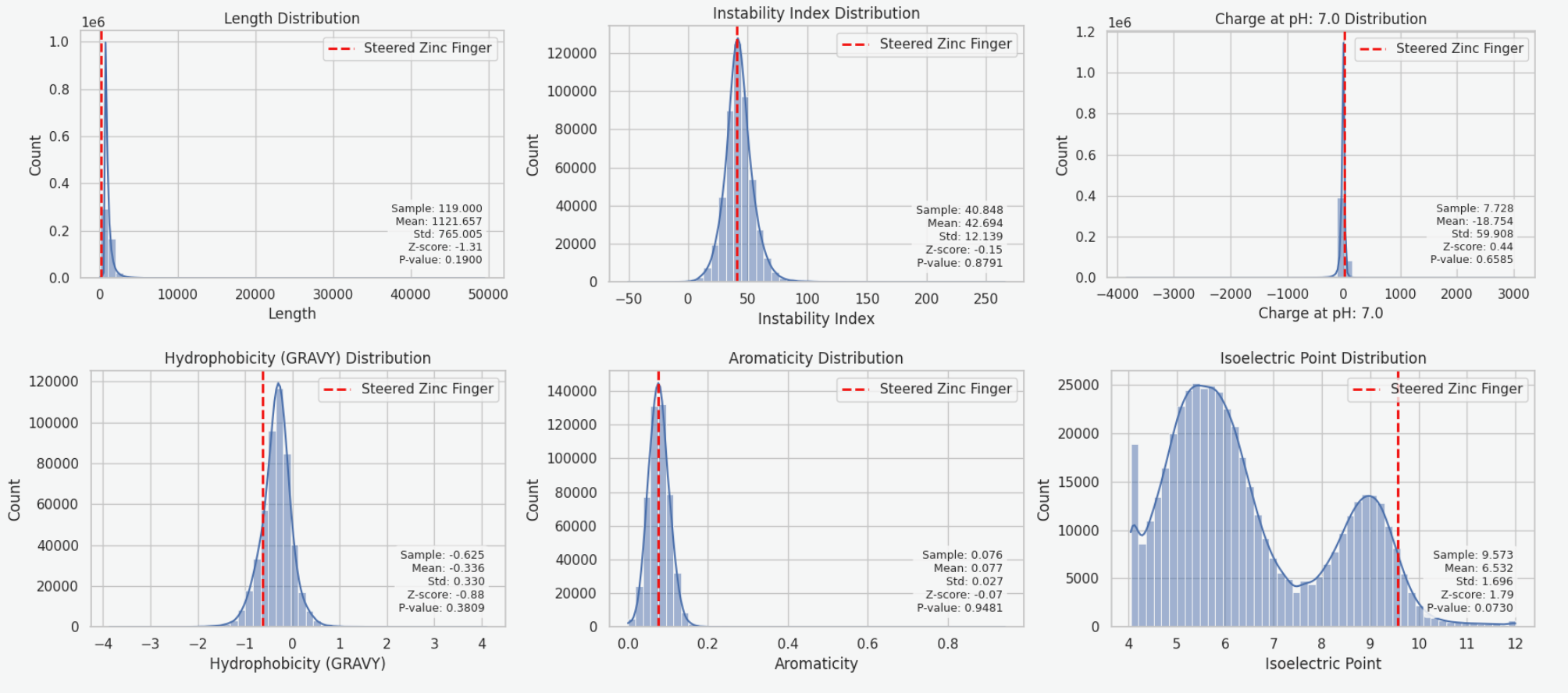}
    \caption{Relative Position of the Generated Zinc Finger Protein}
    \label{fig:zf}
\end{figure}

\begin{figure}[H]
        \centering
        \includegraphics[width=1\linewidth]{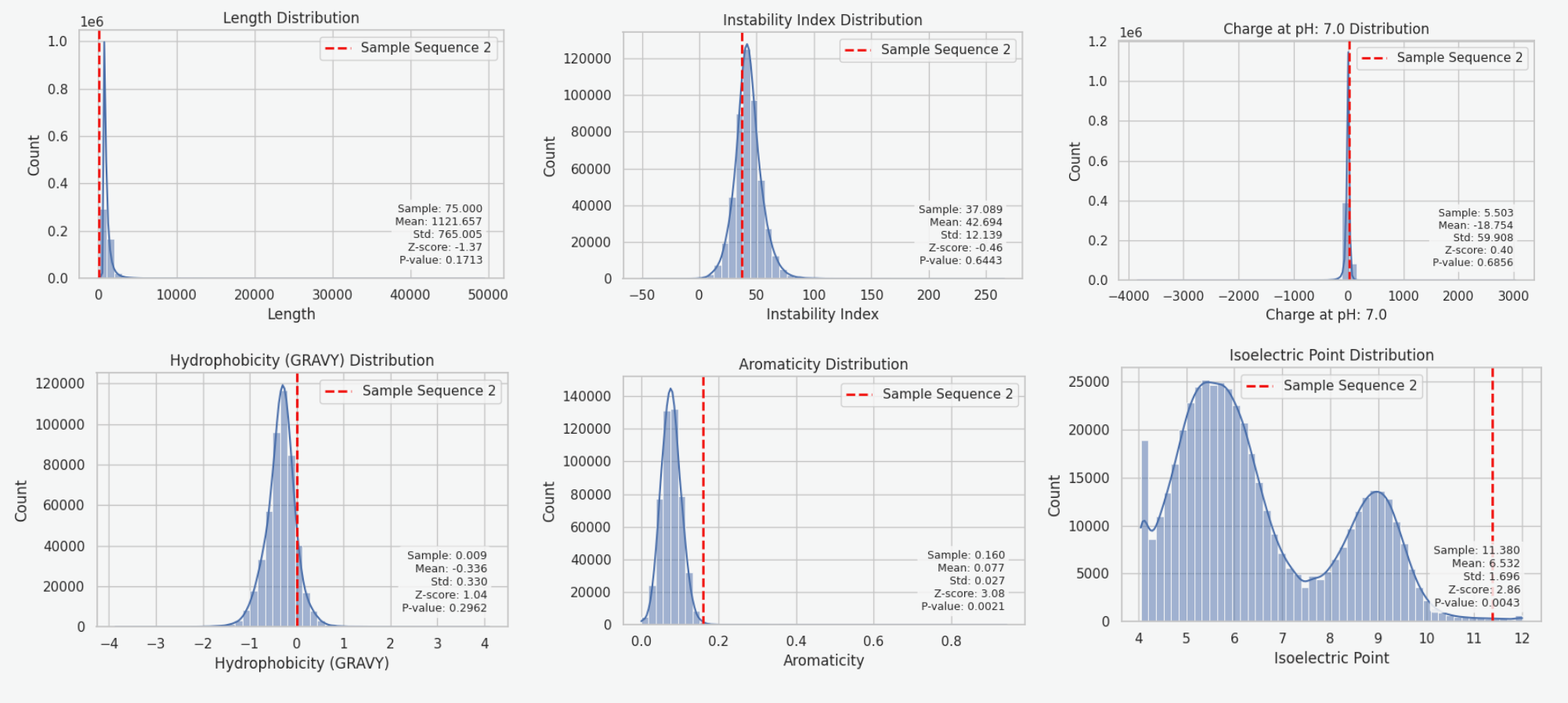}
        \caption{Relative Position of the Generated Alpha Helix Protein}
        \label{fig:alphaanalysis}
\end{figure}

\begin{figure}[H]
        \centering
        \includegraphics[width=1\linewidth]{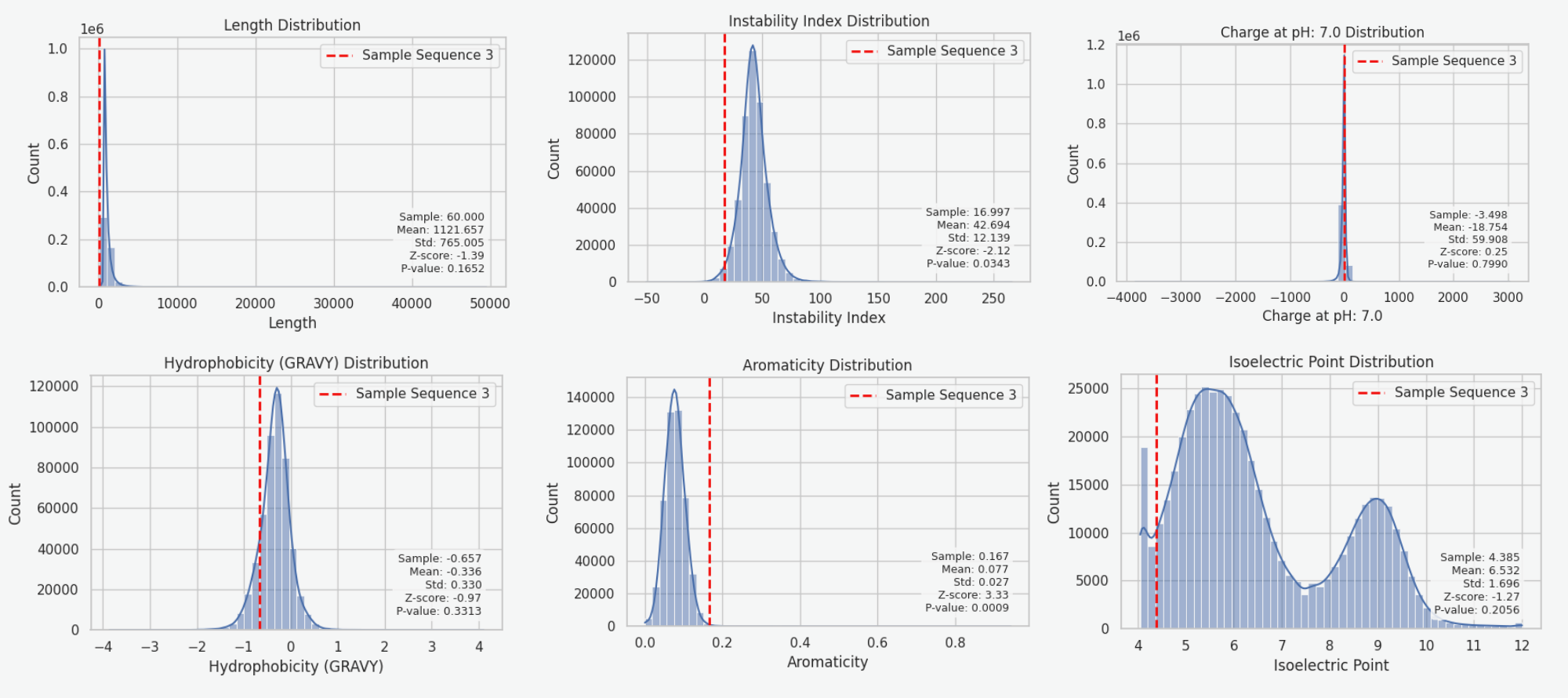}
        \caption{Relative Position of the Generated Beta Strand Protein}
        \label{fig:betaanalysis}
\end{figure}

In general, we find that the characteristics of generated proteins roughly lie in the dataset distribution.

We then assess folding potential by calculating the $\Delta G$ of the various proteins generated. We find that essentially no protein generated would be able to viably fold. The procedure for calculating $\Delta G$ can be found below and is based on existing literature \cite{GUEROIS2002369}:

\begin{lstlisting}
Alpha Protein
BackHbond       =               -20.54
SideHbond       =               -8.28
Energy_VdW      =               -41.05
Electro         =               -2.47
Energy_SolvP    =               64.27
Energy_SolvH    =               -52.90
Energy_vdwclash =               6.10
energy_torsion  =               7.56
backbone_vdwclash=              60.68
Entropy_sidec   =               19.47
Entropy_mainc   =               127.51
water bonds     =               0.00
helix dipole    =               -0.05
loop_entropy    =               0.00
cis_bond        =               0.00
disulfide       =               0.00
kn electrostatic=               0.00
partial covalent interactions = 0.00
Energy_Ionisation =             0.00
Entropy Complex =               0.00
-----------------------------------------------------------
Total          = 				  99.64

\end{lstlisting}
\begin{lstlisting}
Beta Protein

BackHbond       =               -13.22
SideHbond       =               -10.70
Energy_VdW      =               -23.19
Electro         =               -0.83
Energy_SolvP    =               42.47
Energy_SolvH    =               -26.60
Energy_vdwclash =               2.45
energy_torsion  =               11.96
backbone_vdwclash=              39.40
Entropy_sidec   =               14.64
Entropy_mainc   =               116.13
water bonds     =               0.00
helix dipole    =               -0.06
loop_entropy    =               0.00
cis_bond        =               0.00
disulfide       =               0.00
kn electrostatic=               0.00
partial covalent interactions = 0.00
Energy_Ionisation =             0.00
Entropy Complex =               0.00
-----------------------------------------------------------
Total          = 				  113.05
\end{lstlisting}

\begin{lstlisting}
Zinc Finger
BackHbond       =               -37.56
SideHbond       =               -21.78
Energy_VdW      =               -74.44
Electro         =               -1.94
Energy_SolvP    =               111.03
Energy_SolvH    =               -96.64
Energy_vdwclash =               5.35
energy_torsion  =               15.02
backbone_vdwclash=              80.42
Entropy_sidec   =               43.27
Entropy_mainc   =               230.07
water bonds     =               0.00
helix dipole    =               -0.04
loop_entropy    =               0.00
cis_bond        =               0.00
disulfide       =               0.00
kn electrostatic=               0.00
partial covalent interactions = 0.00
Energy_Ionisation =             0.15
Entropy Complex =               0.00
-----------------------------------------------------------
Total          = 				  172.49

\end{lstlisting}

\newpage


\end{document}